\documentclass[10pt, a4paper]{article}

\usepackage[final]{lrec2026} %
\usepackage{microtype}         %
\usepackage{graphicx}          %
\usepackage{url}               %
\usepackage{hyperref}          %
\usepackage{natbib}            %
\usepackage{enumitem}          %
\usepackage{booktabs}          %
\usepackage{arydshln}          %
\usepackage{nicematrix}        %
\usepackage[capitalise]{cleveref} %
\usepackage{xcolor} %
\usepackage{float}
\usepackage{amsmath}
\usepackage{amssymb}  %
\usepackage{natbib}

\title{Low-Rank Compression of Language Models \\ via Differentiable Rank Selection}

\name{Sidhant Sundrani \quad Francesco Tudisco \quad Pasquale Minervini} 

\address{
University of Edinburgh, Edinburgh, UK \\
sidhantls@outlook.com \quad \{f.tudisco, p.minervini\}@ed.ac.uk
}

\abstract{
Approaches for compressing large-language models using low-rank decomposition have made strides, particularly with the introduction of activation and loss-aware SVD, which improves the trade-off between decomposition rank and downstream task performance.
Despite these advancements, a persistent challenge remains--selecting the optimal ranks for each layer to jointly optimise compression rate and downstream task accuracy.
Current methods either rely on heuristics that can yield sub-optimal results due to their limited discrete search space or are gradient-based but are not as performant as heuristic approaches without post-compression fine-tuning.  
To address these issues, we propose Learning to Low-Rank Compress (LLRC), a gradient-based approach which directly learns the weights of masks that select singular values in a fine-tuning-free setting.
Using a calibration dataset, we train only the mask weights to select fewer and fewer singular values while minimising the divergence of intermediate activations from the original model.
Our approach outperforms competing ranking selection methods that similarly require no post-compression fine-tuning across various compression rates on common-sense reasoning and open-domain question-answering tasks.
For instance, with a compression rate of 20\% on Llama-2-13B, LLRC outperforms the competitive Sensitivity-based Truncation Rank Searching (STRS) on MMLU, BoolQ, and OpenbookQA by 12\%, 3.5\%, and 4.4\%, respectively. Compared to other compression techniques, our approach consistently outperforms fine-tuning-free variants of SVD-LLM and LLM-Pruner across datasets and compression rates. Our fine-tuning-free approach also performs competitively with the fine-tuning variant of LLM-Pruner.
 \\ \newline  }

\begin{document}
\pagestyle{plain}

\maketitleabstract

\section{Introduction}
Large language models (LLMs) such as GPT-3~\citep{gpt3} and LLaMA~\citep{llama} show remarkable results in natural language understanding and generation tasks. %
These models are not just pivotal in zero-shot language modelling but also extend their utility to applications such as code generation~\citep{codex}, conversational agents~\citep{llama_chatbot}, and personalised education~\citep{chatgpt_education}.
Despite the success of these models in solving a wide range of tasks, their use is limited by high computing and memory requirements. For example, LLaMA-2 comes in 7 billion and 40 billion parameter variants~\citep{llama}, requiring 25.79 GB and 153.87 GB of memory, respectively. As these models grow, various compression techniques have been developed to reduce their size.  

Quantisation, which reduces the number of bits required to represent each parameter, is widely used for compressing language models~\citep{dettmers,huggingface_quant,awq}.
As an alternative to quantisation, other works explored the structural pruning of LLMs~\citep{llm_pruner, sheared_llama}. 
These works follow two stages for compression: first, pruning for model compression and then a training stage to recover performance \citep{llm_pruner,sheared_llama}. 
  
In addition to quantisation and pruning, recent works show that applying low-rank matrix decomposition techniques can also compress and reduce memory requirements of language models~\citep{weighted_svd,losparse,asvd, svdllm, a3}.

Recent methods have further refined low-rank compression through weighted or activation-aware singular value decomposition, a variant that makes the decomposition loss-aware or activation-aware~\citep{weighted_svd,asvd, adaptive, a3}. Nevertheless, a persistent challenge across these low-rank approaches is the selection of optimal layer-wise ranks, which ultimately governs the achievable balance between parameter reduction and model fidelity.

Given that research has shown that different layers of a language model may have different optimal compression rates~\citep{asvd,dmc}, using a constant rank across layers may not be an effective solution.
To address this problem, \citet{asvd} proposes a heuristic named Sensitivity-based Truncation Rank Searching (STRS), which iteratively searches for optimal ranks per layer by evaluating model perplexity on a small calibration set.
Despite proving better than a naive selection of constant compression ratios across layers, this approach has two core problems that can lead to suboptimal solutions.
First, the search space of ranks is a discrete set of only 10 elements, significantly restricting the number of options available for the ranks.
Second, the optimal decomposition of each layer is identified independently, without taking into account the decomposition of the other layers. %

\begin{figure*}[t]
    \vspace{-8mm} 
    \centering
    \includegraphics[width=\linewidth]{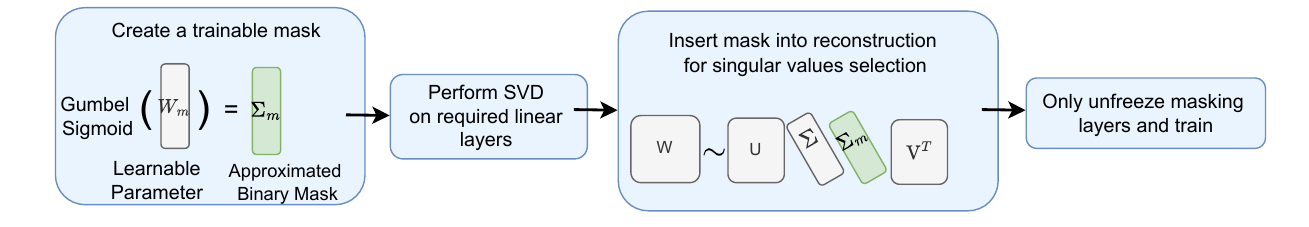}
    \caption{Outline of our proposed method to learn SVD ranks for low-rank compression}
    \label{fig:outline}
    \vspace{-5mm} 
\end{figure*}

On the other hand, \citep{adaptive} proposed a rank selection approach to learn the optimal ranks through gradient descent. In Adaptive Rank Selection (ARS), a binary masking mechanism is used for optimising the number of ranks through training \citep{adaptive}. 
Using GRUs \citep{gru} and linear projections, ARS introduced a learnable singular value masking layer into the SVD reconstruction from which the rank was extracted.
However, a key shortcoming is that rank selection using ARS leads to heavy performance degradation and requires an expensive post-compression training stage. Moreover, the work does not explore optimal methods to convert the learnt mask into a final low-rank model. Similar to this approach, pruning techniques as \citep{structured_pruning} have also explored learning singular value masking for compression; however, unlike ARS, it performs training of the entire model and focuses on smaller models like BERT.
To address these problems in rank selection in low-rank decomposed models, we propose a method called \emph{Learning to Low-Rank Compress} (LLRC).

LLRC can learn the optimal per-layer factorisation ranks by introducing a singular value selection mask $\Sigma_{m}$ into the matrix reconstruction, which is optimised via gradient-based optimisation; \cref{fig:outline} provides a high-level outline of the method. The mask $\Sigma_{m}$ is trained using a multi-objective loss function that enables the balancing of compression costs and downstream task performance. This training approach is lightweight, as it only requires gradient computation for the linear singular value masking layers rather than for the entire model. Following training, after the overall desired compression rate is achieved, the masks are directly utilised to select the most optimal number of singular values. We also define an optimal approach to handle effectively uncompressed layers after training to improve performance. To summarise, our key contributions are the following:

\begin{itemize}[leftmargin=10pt,noitemsep,nolistsep]
    \item A fine-tuning free technique for LLMs called \textit{Learning to Low-Rank Compress (LLRC)} to learn optimal singular values for each layer through training on a small calibration dataset.
    \item A learnable singular value masking linear parameter that learns, in a fine-tuning-free setting, to select the most optimal \textit{any-k} singular values for compression of LLMs.
\end{itemize}

\section{Related Work}
Model compression is a crucial field in deep learning that focuses on reducing the computational costs associated with deploying models while maintaining their performance.
There are several approaches to neural network compression, including pruning~\citep{pruning_1,pruning_2,llm_pruner}, quantisation~\citep{quantization_1, quantization_2}, and low-rank factorisation, the focus of this work.
Specifically in natural language processing (NLP), there have been various efforts along these lines.
Early work aimed to reduce the number of parameters in LSTMs.
For instance, \citep{rnn_svd} applied low-rank decomposition techniques such as Semi-NMF and SVD for LSTM compression and compared the results to pruning.
In contrast to performing low-rank factorisation on a trained model, other works applied tensor decomposition to re-parameterise the model architecture for training \citep{tensor_train,tensor_ring,lowrank1}.

More recent works focused on applying low-rank factorisation to compress language models without full re-training \citep{asvd, weighted_svd, svdllm, a3}. For instance, \citep{weighted_svd} developed a weighted singular value decomposition approach called Fisher-Weighted SVD (FWSVD) that preserves the model performance on a given downstream task. Their method outperforms naive SVD \citep{weighted_svd} for compression. Similarly, Activation-Aware SVD (ASVD) incorporates intermediate activations into the decomposition, optimising for output reconstruction rather than weight approximation~\citep{asvd}. Instead of minimizing the output loss of one transformation, A3 aims to minimize the output loss of a group of transformations or components \citep{a3}. They formulate low-rank decompositions across group of weights, to minimize loss in attention scores, attention outputs, and MLP outputs \citep{a3}.

In such approaches, the singular value rank for each layer has to be determined.
Besides a naive approach that selects an equal rank across all layers, recent works explored approaches to find the optimal rank for each layer.
A relevant approach is Sensitivity-based Truncation Rank Searching~\citep[STRS;][]{asvd}, which iteratively searches for optimal ranks per layer by evaluating model perplexity on a small calibration set.
STRS performs a binary search by using a discrete set of pre-defined 10 compression rates and iteratively applying low-rank decomposition to one layer, leaving the rest of the network unchanged~\citep{asvd}.
Whereas, ARS \citep{adaptive} tackle rank selection using gradient-based optimisation, employing a recurrent neural network and linear layers to predict binary masks over singular values. 

In contrast to STRS and ARS, which rely on iterative rank selection procedures, SVD-LLM focused on making the weighted decomposition truncation aware ~\citep{svdllm}. SVD-LLM addresses this by introducing truncation-aware data whitening and establishing a direct correlation between singular value magnitude and layer output error ~\citep{svdllm}.

\section{Background}
We now describe how Singular Value Decomposition (SVD) can be used to compress linear layers.
Moreover, we also briefly cover details of ASVD~\citep{asvd}, a weighted SVD technique which yields higher model compression rates at lower performance loss.

\subsection{Compressing Transformers with SVD}
\label{sec:compressing_svd}
SVD is a fundamental matrix factorisation technique used in various fields, including image processing and machine learning \citep{svd1, svd2}.
The SVD decomposes a matrix \( W \) into the product of three other matrices: \( U_r \), \(\Sigma_r\), and \( V_r^T \) as shown in \cref{eq:svd}:
\begin{equation} \label{eq:svd}
W = U_r \Sigma_r V_r^T
\end{equation}
Here, \( W \in \mathbb{R}^{m \times n} \) is a rank-\(r\) matrix, \( U_r \in \mathbb{R}^{m \times r} \) and \( V_r \in \mathbb{R}^{n \times r} \) are orthogonal, \( \Sigma_r \in \mathbb{R}^{r \times r} \) is the diagonal matrix of the singular values, and \( r \) is the rank of the decomposition.

Decreasing \( r \) reduces the dimension of \( U_r \) and \( V_r \) but worsens the approximation of \( W \). In particular, for any $k\leq r$, it holds:
\[
U_k\Sigma_kV_k^T = \arg\min_{\mathrm{rank}(W_k)=k}\|W_k-W\|_2\, .
\]
This would result in the storage of three smaller matrices, $U$, $\Sigma$, and $V$, %
in place of a larger matrix $W$. %
Therefore, if $W$ is of shape \( (m, n) \), the compression can be quantified by the parameter ratio in \cref{eq:param_ratio}, where $r$ denotes the rank of decomposition:
\begin{equation}
\label{eq:param_ratio}
\text{Param Ratio} = \frac{r(m + n)}{mn}.
\end{equation}

Such an approximation can be applied to linear layers in a transformer, such as the query projections and linear projections in the feed-forward network. 

\subsection{Activation-Aware SVD}
Activation-Aware SVD is a form of weighted SVD that aims to minimise the reconstruction error of the output of a linear transformation rather than minimising the error of the reconstructed weight matrix~\citep{asvd}.
This approach can be formulated as optimising the following quantity:
\begin{equation} \label{eq:asvd}
\arg\min_{\mathrm{rank}(W_k)=k}\left\lVert W_kX - WX \right\rVert_2
\end{equation}
\noindent where $W_k$ is the reconstructed weight using $k$ singular values and $X$ is an input into the linear transformation.
This is achieved by performing SVD on $WS$ rather than $W$ and then scaling the reconstructed weight by $S^{-1}$, where $S$ is a matrix calculated from a set of inputs $X$ designed to capture the influence of the input channels on the weights~\citep{asvd}. 

\section{Learning to Low-Rank Compress}
Learning to Low-Rank Compress (LLRC) centres on applying SVD to each weight matrix targeted for compression, followed by a learnable masking layer that selects singular values, as shown in \cref{fig:outline}.

During training, these masking layers can adaptively learn highly different compression rates for each layer.
Training jointly optimises compression and model performance using a multi-task training objective
The only learnable parameters are the weights that control the masking, while the rest of the model is frozen.
The following sub-sections further provide details of the training procedure.

\subsection{Applying SVD}
We perform SVD on all linear projection layers in the model except for the logits layer.
Given the ability of weighted SVD approaches to retain performance at higher compression rates~\citep{bert_lowrank,asvd}, we utilise ASVD as a drop-in replacement for SVD in most of our experiments.
For consistency with \citep{asvd}, we use $\alpha = 0.5$.

\subsection{Trainable Singular Value Selection}
Following decomposing a linear layer into its factors using SVD as in \cref{eq:svd}, we introduce a learnable mask inserted in its reconstruction to select singular values, outlined in \cref{fig:outline} and is defined as follows:
\begin{equation} 
\label{eq:learnablemask}
W = U \Sigma (\Sigma_{\text{mask}}) V^T,
\Sigma_{\text{mask}} = g(W_{\text{learnable}}).
\end{equation}

The matrix \(W_{\text{learnable}} \in \mathbb{R}^{1 \times \text{rank}}\) is a learnable paramter, used to generate \(\Sigma_{\text{mask}}\). The learnt mask is represented by \(\Sigma_{\text{mask}} \in \{ 0, 1 \}^{1 \times \text{rank}}\) and the reconstructed weight is \(W \in \mathbb{R}^{m \times n}\). The function \(g\) in \cref{eq:learnablemask} represents Gumbel-Sigmoid ~\citep{gumbel}.
Gumbel-Sigmoid re-parameterises the Bernoulli distribution  by injecting stochasticity ~\citep{gumbel} into the learning of the masks $\Sigma_{\text{mask}}$:

\begin{align}
\tilde{b}_t &= \text{sigmoid}\left[\log \frac{\hat{b}_t u}{(1 - \hat{b}_t)(1 - u)}\right]^{1/\tau} \\
u &\sim \text{Uniform}(0, 1)
\end{align}

At the start of training, $W_{\text{learnable}}$ is initialised such that the mask selects all singular values and through training, its parameters learn sparser masks to achieve the target compression rate.
During training, the only learnable parameters in the model are the newly introduced masking layers and the rest of the model is frozen, keeping the training process efficient.

\subsection{Training}
\subsubsection{Distillation Dataset}
On the training corpus, a set of 3000 documents, we create a distillation dataset which contains the hidden state of each token from the middle layer and the hidden states from before the logits layer~\citep{minilmv2}.
The activations in this dataset will serve as labels used during training.

\subsubsection{Optimisation} 
\label{optimization}
The objective function focuses on minimising the number of singular values selected, minimising the divergence between the activations of the original and compressed models, and enforcing the smoothness of the learned masks.
We aim to minimise the total loss \( L \), which is a weighted sum of the compression loss \( L_{\text{compression}} \), the distillation loss \( L_{\text{distillation}} \), and the Total Variation loss $\mathcal{L}_{\text{tv}}$: 
\begin{equation} 
\label{eq:loss_function}
\mathcal{L} = \alpha \mathcal{L}_{\text{distillation}} + \beta \mathcal{L}_{\text{compression}} + \gamma \mathcal{L}_{\text{tv}},
\end{equation}
\noindent where $\alpha, \beta, \gamma \in \mathbb{R}$ are hyper-parameters. These parameters are selected to guide compression rates. This is further discussed in \cref{sec:objective_function}.

\paragraph{Compression Loss}
For the compression loss, we directly minimise the mean of the learnable weights, helping generate sparser masks and increasing the compression rate. 
\begin{equation}
\label{eq:compression}
\mathcal{L}_{\text{compression}} = \frac{1}{N_{\text{layers}}} \sum_{i=1}^{N_{\text{layers}}} \text{Average}(W_{\text{learnable},i})
\end{equation}

Through this, the model can learn different compression rates for every layer.

\paragraph{Distillation Loss}
The distillation loss minimises the differences in activations between the original model and the model being compressed using the mean-squared error loss in \cref{eq:disillation}.
We use the following distillation loss:
\begin{equation}
\mathcal{L}_{\text{distillation}} = \| A_{\text{compressed}} - A \|_F^2,
\label{eq:disillation}
\end{equation}
\noindent where \( A_{\text{compressed}} \in \mathbb{R}^{L\times D} \) and \( A \in \mathbb{R}^{L\times D} \) denote the activations of the compressed and original models, respectively, where \( L \) is the sequence length and \( D \) is the dimension of the hidden state.
Following \citep{minilmv2}, we use the activations of the middle layer and the hidden states before the logits layer as targets.

\paragraph{Total Variation Loss}
As described in \cref{eq:tv_loss}, we introduce a Total Variation (TV) loss on the learnable mask from \cref{eq:learnablemask} to guide learning a smooth mask. Here, $n$ refers to the index of the mask vector of length $N$. A smooth mask refers to one that selects neighbouring singular values together rather than skipping intermediate values.
\begin{equation}
\mathcal{L}_{\text{tv}} = \sum_{n=0}^{N-1} \left| \Sigma_{\text{mask}, n+1} - \Sigma_{\text{mask}, n} \right|
\label{eq:tv_loss}
\end{equation}
Theoretically, reconstructing a weight matrix after SVD involves utilizing the top-k largest singular values sorted by their magnitude.
Following this intuition, if, for example, the 5th and 7th singular values are deemed relevant by the mask, this loss forces the 6th singular value also to be deemed relevant by the mask.
We experimentally analyse the contribution of the TV loss in our ablations in \cref{sec:tv_loss_ablation}.

\subsection{Model Post-Processing}
\label{sec:post}

After training, we use the learned singular value masks to select the required singular values for compression.
To obtain the binary mask, we applied a 0.5 threshold to the sigmoid of the logits, without needing the stochastic masking operator.

If a layer is negligibly compressed, we reconstruct its entire weight matrix using its full rank, and the layer remains uncompressed.
This design choice of avoiding using the mask is crucial to maintain performance; even if the learned singular value mask retains 90\% of the singular values, compression might not be achieved as per \cref{eq:param_ratio}, and using the full rank will better preserve performance. Empirically, we find this heuristic highly effective for performance and is explored in Ablation \cref{sec:model_conversion_ablation}.
  
After training, for compressed weights, we retain only the left and right singular vectors, merging the singular values into one of them for computational efficiency.

\section{Experimental Setup}
\begin{figure*}[h!]
\begin{tabular}{ll}
 \rotatebox{90}{$\quad$Llama-2-7B} &   \includegraphics[width=.90\textwidth,clip, trim=6em 6em 6em 6em]{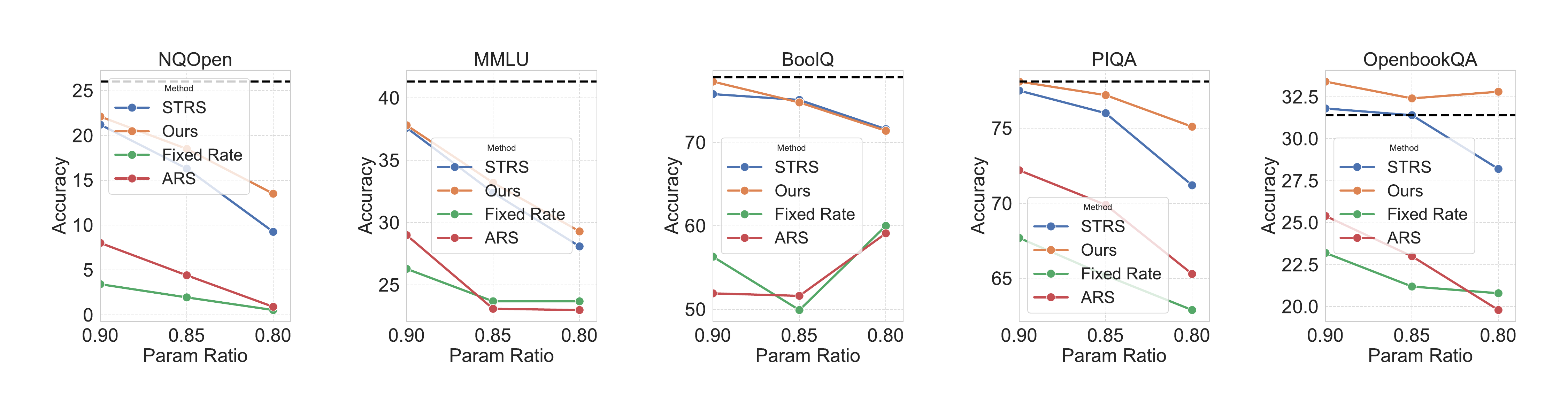}\\
   \rotatebox{90}{$\quad$Llama-2-13B} &   \includegraphics[width=.90\textwidth,clip, trim=6em 6em 6em 6em]{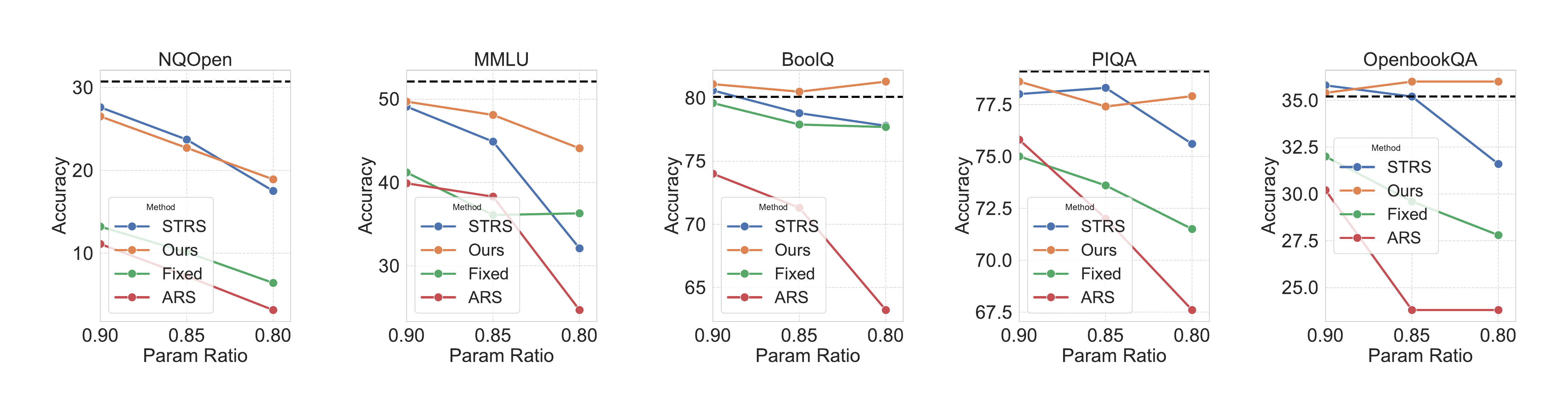} \\
\rotatebox{90}{$\qquad$Gemma-7B} &   \includegraphics[width=.90\textwidth,clip, trim=6em 6em 6em 6em]{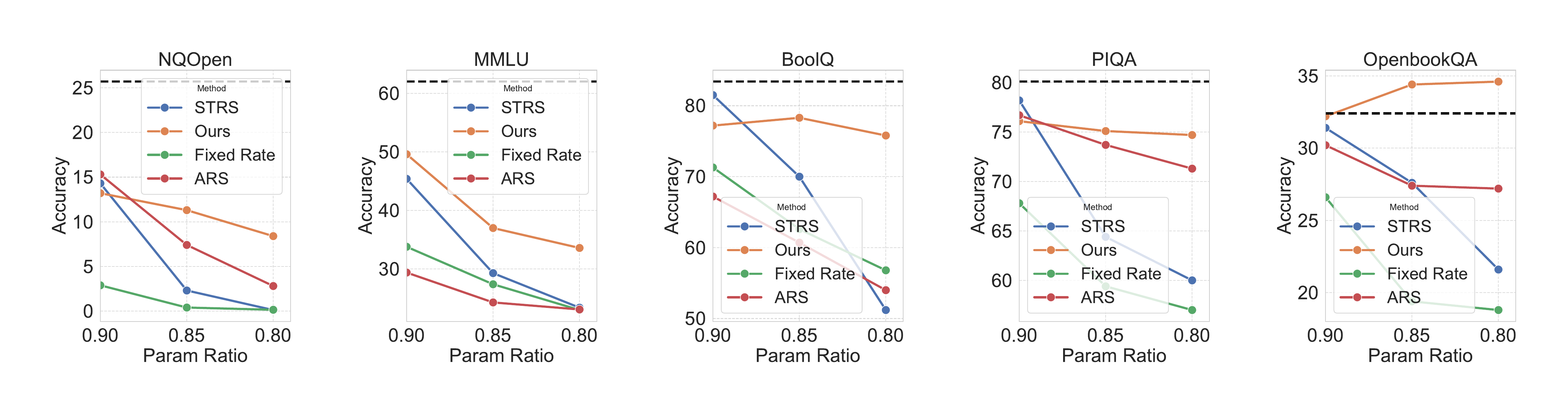}\\
\rotatebox{90}{$\qquad$Llama-3-8B} &   \includegraphics[width=.90\textwidth,clip, trim=6em 6em 6em 6em]{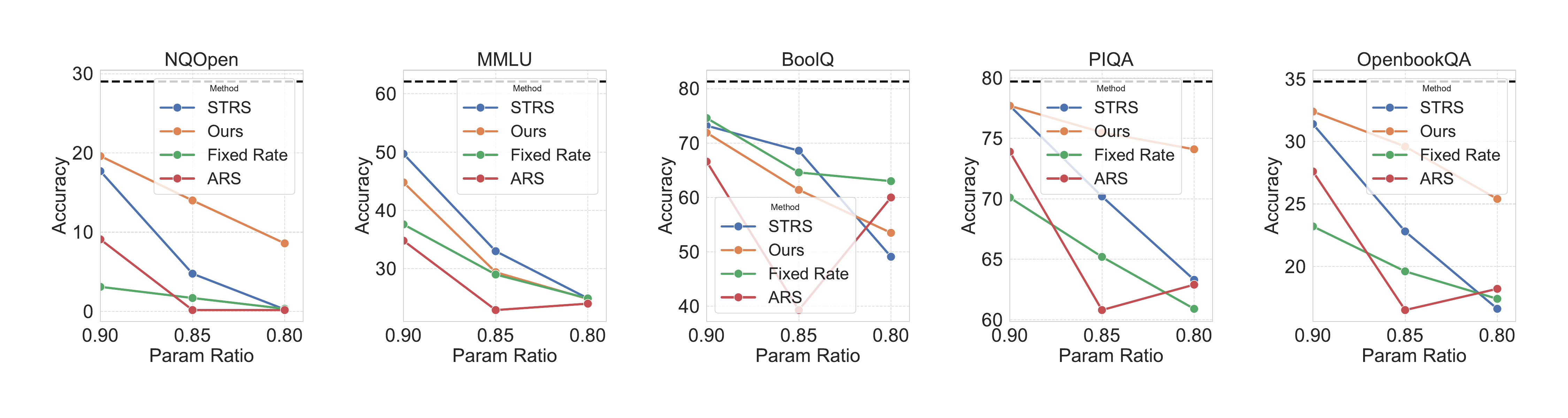}\\
\end{tabular}
\caption{Model performance using different rank selection methods: Fixed Rate (naive baseline), STRS (Yuan et al., 2024), ARS (Gao et al., 2024), and our proposed approach.}
\label{fig:metrics_together}
\end{figure*}

\subsection{Evaluation}
We evaluate the compressed model on zero-shot PIQA~\citep{piqa}, BoolQ~\citep{boolq}, OpenbookQA~\citep{openbookqa}, MMLU~\citep{mmlu} and 5-shot evaluations on NQ-Open~\citep{nqopen}.
These are English datasets that cover common-sense reasoning and open-domain QA.

\subsection{Training and Data Configuration} 
\label{distillation_exp}
The training dataset for LLRC consisted of 3,000 unique documents from WikiText-2 ~\citep{wikitext}.
To have a greater number of token activations present per batch, all documents selected had more than 150 words. Experiments were conducted using a batch size of 4, a maximum token length of 256, and optimised with the AdamW optimiser~\citep{adamw}.
To avoid unnecessary training, we use early stopping, terminating the training 750 steps after the target parameter ratio is achieved.

\subsection{Masking Layer Initialization}
The learnable weight matrix \( W_{\text{learnable}} \) is initialised such that the model starts in an uncompressed state, selecting all singular values.
Given that larger singular values are theoretically more significant, we introduce an inductive bias into the weight initialisation of the mask.
To this end, \( W_{\text{learnable}} \) is initialised with linearly distributed values ranging in $[3, 6]$, aligned with the magnitude of the singular values.
Following \citep{dmc}, we use a temperature of 0.1 for the Gumbel function.

\subsection{Objective Function Configuration}
\label{sec:objective_function}
The objective function in \cref{eq:loss_function} represents a weighted sum of the compression, distillation, and total variation loss. The total variation loss weight, $\gamma$, is a constant 1. 
The weight on the compression loss, denoted by $\beta$, is set to a constant value of 1 until the target compression ratio is achieved, after which it is set to 0 to prevent further unnecessary compression and focus on the model loss.
Instead of using a constant value for $\alpha$, which is more susceptible to the initial choice, inspired by \citep{alpha}, we oscillate $\alpha$ between two bounds, 1 and 0, using the cosine function.
This also enables the training to oscillate focus between optimising for compression and performance.

\section{Results}

\subsection{Evaluation Benchmarks}

\begin{table*}[h!]
\centering
\setlength\tabcolsep{2pt}
\small
\resizebox{\textwidth}{!}{
\begin{NiceTabular}{lccccccccccc}
\toprule
\textbf{Method} & \textbf{Param Ratio} & \multicolumn{5}{c}{\textbf{Llama-2-7b}} & \multicolumn{5}{c}{\textbf{Llama-2-13b}} \\
\cmidrule(lr){3-7} \cmidrule(lr){8-12}
 &  & \textbf{NQ-Open} & \textbf{MMLU} & \textbf{BoolQ} & \textbf{PIQA} & \textbf{OQA} & \textbf{NQ-Open} & \textbf{MMLU} & \textbf{BoolQ} & \textbf{PIQA} & \textbf{OQA} \\
\midrule
Baseline & 1.00 & 26.0 & 41.3 & 77.8 & 78.1 & 31.4 & 30.7 & 52.1 & 80.1 & 79.1 & 35.2 \\
\midrule
Pruner & 0.90 & 15.5 & 28.5 & 64.8 & 77.3 & 31.0 & 21.2 & 45.4 & 73.4 & \underline{79.4} & 35.2 \\
SVD-LLM (w) & 0.90 & 10.8 & 24.0 & 49.0 & 69.4 & 28.2 & 18.3 & 31.6 & 76.6 & 73.9 & 30.6 \\
Ours & 0.90 & \underline{22.1} & \underline{37.8} & \underline{77.3} & \underline{78.1} & \underline{33.4} & \underline{26.5} & \underline{49.7} & \underline{81.1} & 78.6 & \underline{35.4} \\
\midrule
Pruner & 0.80 & 7.8 & 25.1 & 64.6 & \underline{76.2} & 29.0 &  12.6 & 23.2 & 67.2 & \underline{78.3} & 32.2 \\
SVD-LLM (w) & 0.80 & 5.87 & 23.8 & 44.4 & 66.1 & 25.6 & 12.9 & 29.6 & 73.9 & 69.8 & 28.8 \\
Ours & 0.80 & \underline{13.5} & \underline{29.3} & \underline{71.4} & 75.1 & \underline{32.8} & \underline{18.9} & \underline{44.1} & \underline{81.3} & 77.9 & \underline{36.0} \\
\midrule
\multicolumn{11}{l}{Methods with additional fine-tuning on Alpaca dataset}\\
\midrule
{Pruner+Finetune} & 0.90 & 17.9 & 34.0 & 71.4 & 78.1 & 33.0 & 22.1 & 48.0 & 76.4 & 79.7 & 36.6 \\
{SVD-LLM+Finetune} & 0.90 & 14.9 & 33.0 & 67.6 & 75.3 & 30.6 & 20.7 & 45.1 & 80.0 & 77.6 & 33.8 \\

\hdottedline
{Pruner+Finetune} & 0.80 & 11.1 & 26.7 & 67.8 & 77.8 & 30.4 & 16.2 & 32.1 & 72.9 & 79.2 & 35.4 \\
{SVD-LLM+Finetune} & 0.80 & 11.8 & 27.1 & 67.5 & 72.4 & 29.6 & 17.6 & 41.3 & 78.6 & 76.3 & 32.4 \\
\bottomrule
\end{NiceTabular}
}
\caption{Performance comparison between LLM-Pruner \citep{llm_pruner}, SVD-LLM \citep{svdllm}, and our approach on LLAMA-2-7B and LLAMA-2-13B.}
\label{tab:combined_metrics}
\end{table*}

To evaluate the efficacy of our gradient-based rank selection  procedure, we benchmark it against a baseline rank selection, Sensitivity-based Truncation Rank Searching~\citep[STRS;][]{asvd}, and Adaptive Rank Selection~\citep[ARS;][]{adaptive}.
We perform extensive compression performance comparisons on four architectures of different sizes: Llama-2-7B, Llama-3-8B, Gemma-7B, and Llama-2-13B.
As a baseline algorithm, `Fixed Rate', the rank is selected to compress each layer equally to the target parameter ratio. To compare fine-tuning free-rank selection approaches, for ARS, we only use the rank selection portion of the algorithm without the post-training fine-tuning. We performed ARS using $\lambda$=16 and $\gamma$=1 \citep{adaptive} on the same dataset we used. Further details of hyperparameters of these approaches can be found in Appendix \cref{sec:hparams_eval}. 

As shown in \cref{fig:metrics_together}, our approach consistently achieves higher accuracy across most datasets, particularly at lower compression ratios. At a parameter ratio of 0.80, our method outperforms STRS on Llama-2-7B by 4.3\% (NQopen), 3.9\% (PIQA), and 4.6\% (OpenbookQA). On Llama-2-13B, improvements reach 12\% on MMLU, 3.5\% on BoolQ, and 4.4\% on OpenbookQA.%

The competitive performance of our approach is rooted in several ablation studies on how to handle learning singular value masks. For instance, as described in Ablations \cref{sec:model_conversion_ablation}, we find that our post-training heuristic of ignoring the learned masks for trivially compressed layers is effective in retaining performance. The numerical results corresponding to \cref{fig:metrics_together} are tabulated in Appendix \cref{sec:all_metrics_appendix}.

\begin{figure*}[htbp]
  \centering
  \includegraphics[scale=0.4]{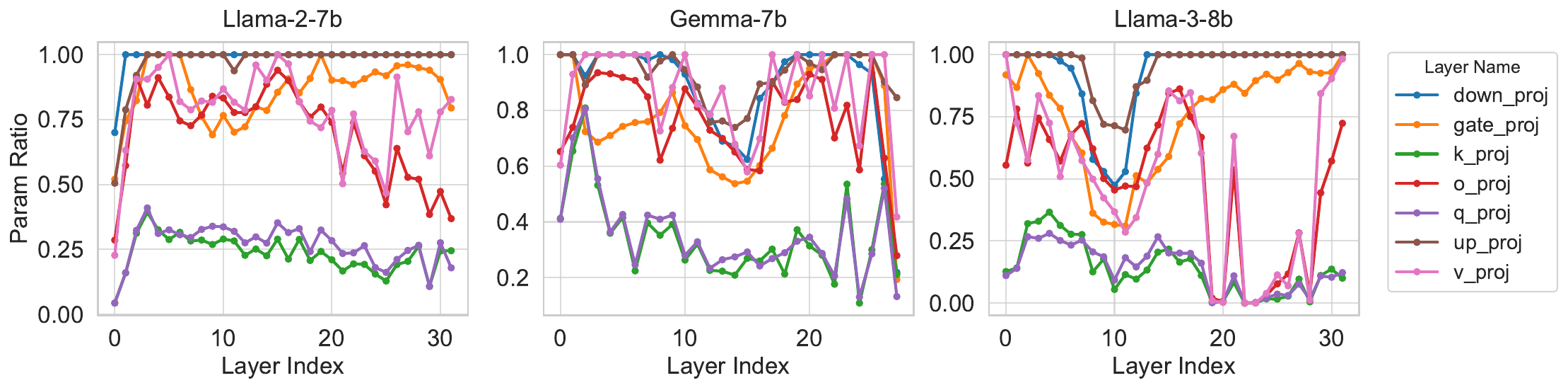}
  \caption{Distribution of compression rates across layer types and layer numbers in models that were compressed by 20\% (param ratio of 0.80) using our approach.}
  \label{fig:compression_ratio}
\end{figure*}

\subsection{Impact of Fine-Tuning}
To place our work in the broader context of model compression, we benchmark our approach against LLM-Pruner and SVD-LLM. LLM-Pruner~\citep{llm_pruner} is a structured pruning method that supports both fine-tuning and fine-tuning-free variants. SVD-LLM is a truncation-aware low-rank compression method, which also relies on post-compression fine-tuning for improved performance recovery ~\citep{svdllm}. Another relevant method is A3~\citep{a3}; however, we could not include it in our benchmarks as it was recently released and lacks a public implementation.

We conduct experiments on Llama-2-7B, Llama-2-13B, and Llama-3-8B. Because the publicly available SVD-LLM implementation does not support Llama-3-8B, our comparisons with LLM-Pruner on this model is present in Appendix \cref{sec:llm_pruner_llama3}. Since our approach does not update model weights, we focus primarily on comparing against the fine-tuning-free variants of SVD-LLM and LLM-Pruner. Nevertheless, we also report results for their fine-tuning variants and observe that our method remains competitive even when those methods leverage additional fine-tuning. The hyperparameters used in these experiments can be found in Appendix \cref{sec:hparams_eval}. 

The comparative analysis in \cref{tab:combined_metrics} demonstrates that our method significantly outperforms the fine-tuning-free variants of LLM-Pruner and SVD-LLM on both Llama-2-7B and Llama-2-13B. For example, at a parameter ratio of 0.80 on Llama-2-13B, our approach surpasses SVD-LLM (w) on all evaluated datasets and outperforms LLM-Pruner on 4 out of 5 datasets. This demonstrates that our rank selection approach better preserves performance in a fine-tuning-free manner. 

In the fine-tuning setting, our approach—despite requiring no fine-tuning—consistently outperforms SVD-LLM even when SVD-LLM is fine-tuned, and remains competitive with the fine-tuned variant of LLM-Pruner. At a parameter ratio of 0.80 on Llama-2-13B, our method exceeds the performance of fine-tuned LLM-Pruner on 4 out of 5 datasets. However, on Llama-3-8B, LLM-Pruner with fine-tuning achieves higher accuracy than our method, as shown in \cref{tab:combined_metrics_llama3}.

\subsection{Insights into Learnt Compression Rates}

To understand the robustness of our rank selection method, we analyze how our training procedure discovers optimal compression strategies. To do this, we examine the learned singular value masks across different models after training. As visualized in \cref{fig:compression_ratio}, the learned compression rates demonstrate that optimization captures the varying compressibility of layers: earlier layers, along with key and query projection layers, consistently exhibit the highest compression rates across all models, while feedforward layers (up/down projections) remain largely uncompressed.

\section{Ablations}
\subsection{Selecting Any-k Singular Values Vs Top-k}
\label{sec:topk_vs_anyk}
Theoretically, when selecting \(k\) singular values for retention, the optimal choice to minimise reconstruction loss would typically be the top \(k\) singular values by magnitude. However, our approach, which allows the mask to learn to select any \(k\) singular values rather than strictly the top \(k\), has shown strong practical performance despite limited theoretical backing.

\begin{figure}[htbp]
  \centering
  \includegraphics[width=0.4\textwidth,clip, trim=.5em 0em .5em 0em]{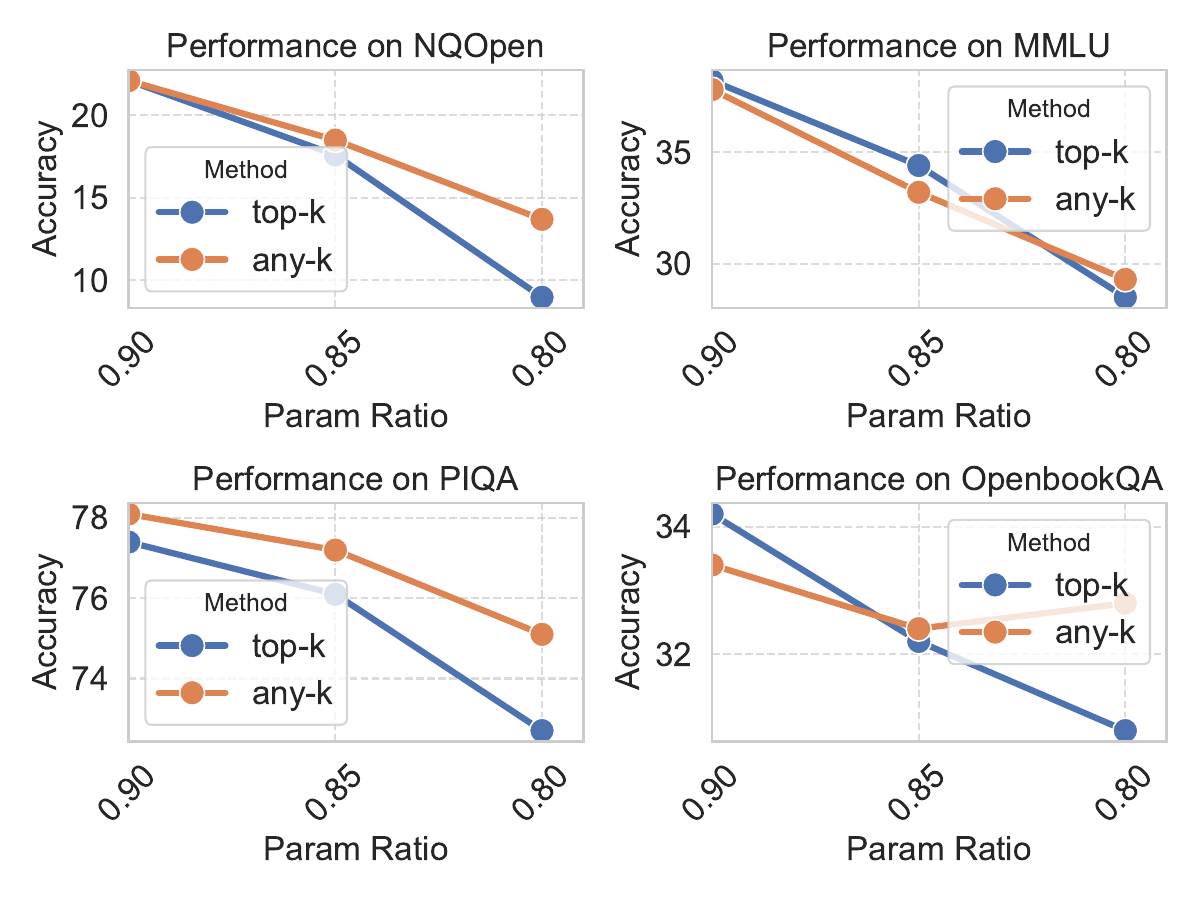}
  \caption{Evaluation performance of Llama-2-7b with using any-k and top-k masking}
  \label{fig:topk_vs_any}
\end{figure}
We compressed the LLaMA-2-7b model and performed evaluation using both the top-\(k\) and any-\(k\) singular value selection. Results in Figure \ref{fig:topk_vs_any} demonstrate that in the lowest parameter ratio of 0.80, any-\(k\) mask outperforms the top-\(k\) mask on all datasets. Significant gains are seen in open-domain QA with any-k performing ~4\% better at parameter ratio 0.80.

\subsection{Introduction of Total Variation Loss}
\label{sec:tv_loss_ablation}
Motivated by the theoretical understanding that higher singular values are more relevant, we used the Total Variation (TV) loss to guide our learnt masks to be smooth.  

\begin{figure}[htbp]
  \centering
  \includegraphics[width=0.4\textwidth,clip, trim=.5em 0em .5em 0em]{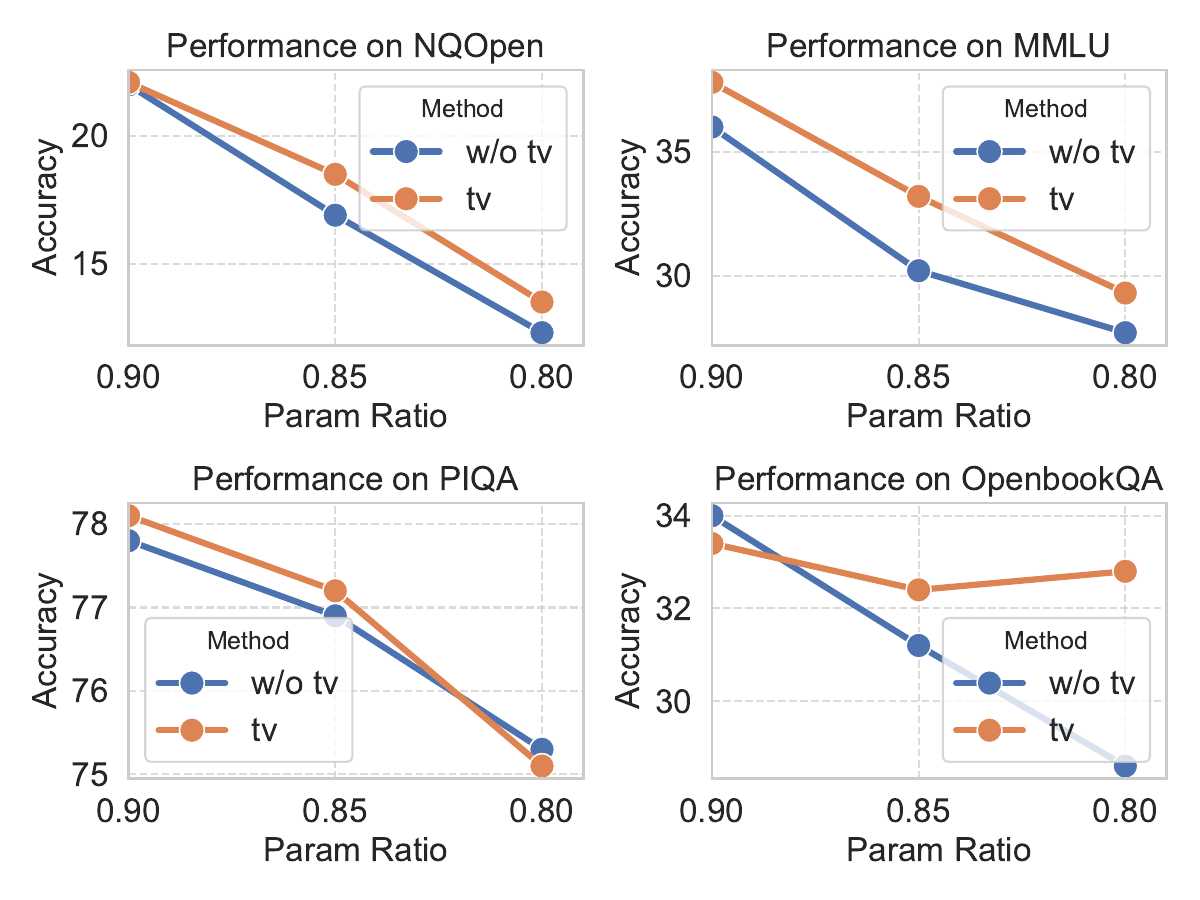}
  \vspace{-1em}
  \caption{Evaluation performance of Llama-2-7b with and without using total variational loss}
  \label{fig:tv_vs_no_tv}
\end{figure}

To validate this hypothesis, we compressed the LLaMA-2-7b model to various target parameter ratios, using and without using the total variational loss. Results in Figure \ref{fig:tv_vs_no_tv} demonstrate the efficacy of introducing the total variational loss. On NQ-Open, MMLU and PIQA introducing this loss function consistently leads to higher performance. On OpenbookQA, this loss function leads to higher performance on parameter ratios of 0.85 and 0.80. 

\subsection{Post-Training Model Conversion}
\label{sec:model_conversion_ablation}
\begin{figure}[htbp]
  \centering
\includegraphics[width=0.4\textwidth,clip, trim=.5em 0em .5em 0em]{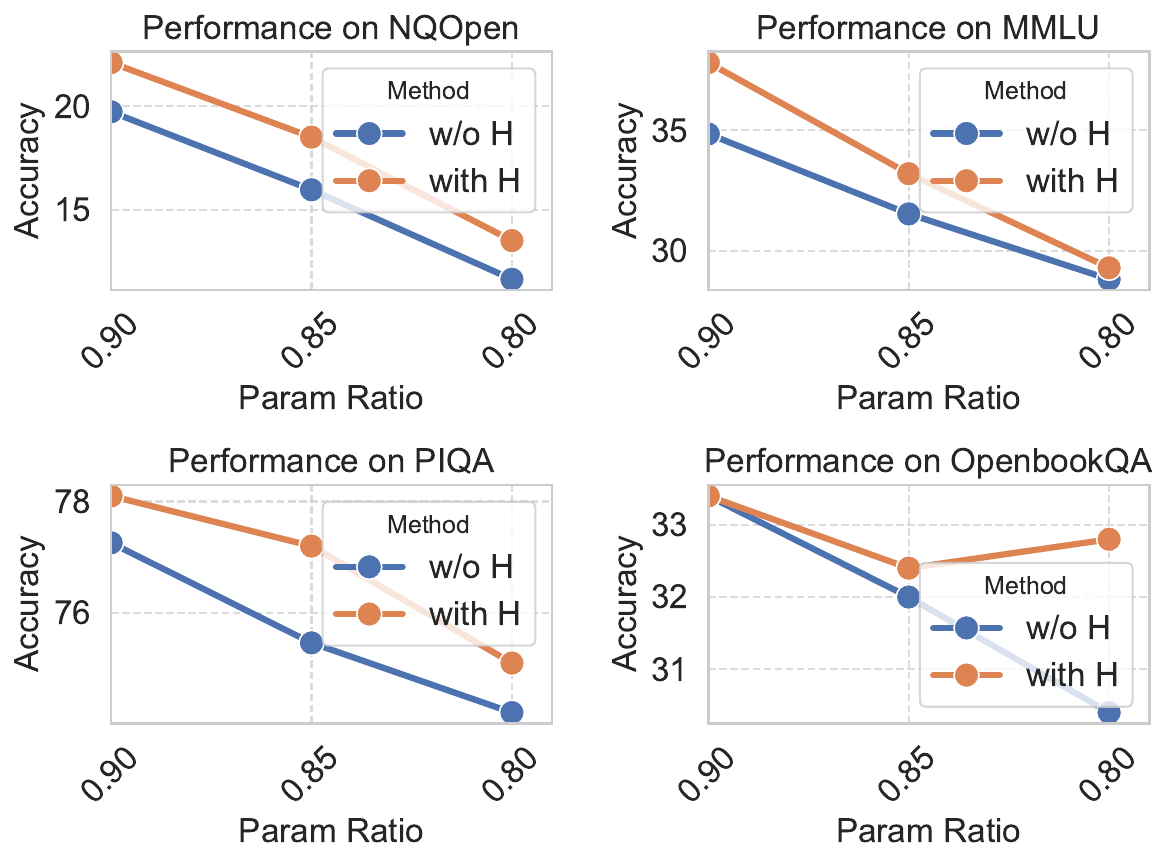}
  \caption{Performance comparison with and without our heuristic for compression: Llama-2-7b}
  \label{fig:heur_small}
\end{figure}

As outlined in \cref{sec:post}, we finalise model compression by retaining only the singular vectors selected by the mask. However, as implied by \cref{eq:param_ratio}, even if the learnt layer drops singular values, it does not always lead to compression. In cases where layers remain effectively uncompressed, we find it highly effective to disregard the learned masks for these layers and instead reconstruct the original matrix as a standard linear layer. This approach is not only theoretically justified but also empirically validated, as illustrated in \cref{fig:heur_small}. 

By applying this heuristic, our method consistently outperforms baseline approaches on nearly all downstream tasks. To the best of our knowledge, this has been unexplored in prior gradient-based rank selection compression methods.

\section{Conclusion}
We introduced Learning to Low-Rank Compress (LLRC), a gradient-based compression method that learns optimal SVD ranks from a small calibration set. By casting rank selection as a learnable mask vector, LLRC trains more efficiently than search procedures that rely on recurrent neural networks (e.g., ARS) and, unlike most alternatives, retains \textit{any-k} rather than just the top-k singular values. Empirically, LLRC consistently outperforms all prior rank-selection approaches across Llama-2-7B/13B, Llama-3-8B, and Gemma-7B; at 20\% compression on Llama-2-13B, for example, it improves over STRS by 12\% on MMLU and by 4\% on OpenbookQA. Compared to existing compression techniques, our fine-tuning-free method outperforms LLM Pruner and SVD-LLM (without fine-tuning) on various compression rates. Compared to their fine-tuning variants, it outperforms SVD-LLM and remains competitive with LLM-Pruner. Beyond raw performance, the learned parameterized masks also reveal systematic compressibility patterns, which may provide insights for general compression research.

\section{Limitations}
The architecture in our proposed method for compression employs a linear mask to selectively mask singular values during an SVD reconstruction. While this approach simplifies the training process, it introduces some limitations related to memory usage during training. During training, to select singular values, the weight matrices are decomposed into their full rank, which results in a higher number of parameters in each layer compared to the original configuration, as described in Equation \ref{eq:param_ratio}. Additionally, at higher compression rates, such as 20\%, we observe more degradation. We believe this can be a limitation of current SVD approaches (ASVD, Fischer SVD) and their ability to minimise reconstruction loss of the weights with fewer singular values. Exploring the synergy between learnable rank selection and component-wise decomposition techniques, such as A3 \citep{a3}, may therefore be a promising direction for future work.

\appendix
\section{Training Details}
\subsection{Distillation Scale Parameter}
\label{sec:distillation-scale}
The distillation scale parameter $\alpha$ in Equation \ref{eq:loss_function} equations below: 
\begin{equation}
z = \cos\left(\frac{2\pi \cdot 10 \cdot \text{current\_step}}{\text{total\_steps}}\right)
\end{equation}
\begin{equation}
\alpha = \min\left(\max(z, b), c\right)
\end{equation}

This scaling $\alpha$ follows a cosine distribution that completes 1 cycle at 10\% of total training steps, and the min value is capped at b and max c, which are decided based on the rate of compression required. The lower the value of b and c, the faster the model learns the required compression rate. For llama-2-7b, we use b=0.3 and c=1.0, for llama-3-8b we use b=0.25 and c=0.5, for Gemma-7b, we use b=0.5 and c=1.0. As a warmup, for the first 250 steps, we avoid any scaling and use $\alpha=1.0$. 

\section{Evaluation Details}
\label{sec:hparams_eval}

\subsection{ASVD STRS Hyper-Parameters}
The hyperparameters for STRS ASVD are below: 

\begin{itemize}
    \item $\alpha$: 0.5
    \item Number of Calibration Samples: 32 
    \item Max Sequence Length: 2048
\end{itemize}

\subsection{Adaptive Rank Selection (ARS) Hyper-Parameters}
\label{sec:ars-pruner-params}
The benchmarks of our ARS algorithm is based on our implementation, which is open-sourced here: \href{https://github.com/sidhantls/}{GitHub}. The dataset used is the same as one used in our work, Wikitext-2 \citep{wikitext}

\begin{itemize}
    \item \textbf{Llama-2-7b:} $\lambda = 16$, $\gamma = 1$, lr = $1\mathrm{e}{-3}$, Optimizer: Adam
    \item \textbf{Llama-3-8b:} $\lambda = 8$, $\gamma = 2$, lr = $1\mathrm{e}{-3}$, Optimizer: Adam
    \item \textbf{Gemma-7b:} $\lambda = 8$, $\gamma = 2$, lr = $1\mathrm{e}{-3}$, Optimizer: Adam
    \item \textbf{Llama-2-13b:} $\lambda = 16$, $\gamma = 1$, lr = $1\mathrm{e}{-3}$, Optimizer: Adam
\end{itemize}
        
For Llama-2-7b, we used $\lambda$=16 and  $\gamma$=1, instead of $\lambda$=8 and  $\gamma$=2, because the former led to an acceptable NQ-Open performance at Param Ratio 0.90.
            
\subsection{LLM Pruner Hyper-Parameters}
\label{sec:llm-pruner-params}
This section outlines the hyperparameters used for benchmarking LLM-Pruner on downstream datasets with various compression ratios.  

\begin{itemize}
    \item \textbf{Pruning:}
    \begin{itemize}
        \item {Block-wise Pruning:}
        \begin{itemize}
            \item Block MLP Layer Start: 4
            \item Block MLP Layer End: 30
            \item Block Attention Layer Start: 4
            \item Block Attention Layer End: 30
        \end{itemize}
        \item Pruner Type: Taylor
        \item Taylor Strategy: \texttt{param\_first}
    \end{itemize}
    \item \textbf {Post-Training:}
    \begin{itemize}
        \item Dataset: \texttt{yahma/alpaca-cleaned}
        \item LoRA Rank: 8
        \item Number of Epochs: 2
        \item Learning Rate: 1e-4
        \item Batch Size: 64
    \end{itemize}
\end{itemize}

\subsection{SVD-LLM Hyper Parameters}
\label{sec:svd-llm-params}
These are the hyperparameters for the \textbf{SVD-LLM whitening step (without finetuning)}:

\begin{itemize}
    \setlength\itemsep{0em}
    \setlength\parskip{0em}
    \item {whitening\_nsamples}: 256
    \item {dataset}: wikitext2
    \item {seed}: 3
    \item{model\_seq\_len}: 2048
\end{itemize}

As per SVD-LLM \citep{svdllm}, after the whitening step, LoRA fine-tuning is performed separately for the left and right singular values. This is the set of hyperparameters used for each of the two fine-tuning steps. 

\begin{itemize}
    \setlength\itemsep{0em}
    \setlength\parskip{0em}
    \item data\_path: yahma/alpaca-cleaned
    \item lora\_r: 8
    \item num\_epochs: 1
    \item learning\_rate: 1e-4
    \item batch\_size: 64
\end{itemize}

\section{Complexity Analysis of a Low-Rank Linear Layers}
\label{sec:complexity_analysis}
\subsection{Space Complexity}
The compression rate of a low-rank layer is derived in \cref{sec:compressing_svd}. As shown in Equation~\ref{eq:param_ratio}, a weight matrix $\mathbf{W} \in \mathbb{R}^{m \times n}$ that originally requires $mn$ parameters can be represented in low-rank form using only $mr + nr$ parameters for rank $r$. When $r < \min(m, n)/2$, this representation yields a reduction in memory usage.

\subsection{Time Complexity}
When applying SVD-based compression to a weight matrix $W \in \mathbb{R}^{m \times n}$, 
we decompose it into two lower-rank matrices $U \in \mathbb{R}^{m \times r}$ and 
$V \in \mathbb{R}^{r \times n}$, where $r < \min(m,n)$ represents the compressed rank. 
For efficiency, this decomposition is performed after fusing the singular value 
vector into the left eigenvectors.

\textbf{Original FLOPs}: For an uncompressed linear layer with weight matrix $W \in \mathbb{R}^{m \times n}$ and input $X \in \mathbb{R}^{m \times m}$, the matrix multiplication $XW$ requires:
\begin{equation}
\text{FLOPs}_{\text{original}} = 2m^2n
\end{equation}

\textbf{Compressed FLOPs}: After SVD decomposition, the computation becomes $X(UV) = (XU)V$, requiring two sequential matrix multiplications:
\begin{equation}
\text{FLOPS}_{\text{compressed}} = 2m^2r + 2 \cdot mrn
\end{equation}

\textbf{FLOP Ratio}: The computational savings achieved through low-rank decomposition is:

\begin{align}
\text{FLOP Ratio} &= \frac{\text{FLOPs}_{\text{compressed}}}{\text{FLOPs}_{\text{original}}} \notag \\
                  &= \frac{2mr(m + n)}{2m^2n} = \frac{r(m + n)}{mn}
\end{align}

If the FLOP ratio is $< 1$, we can say that fewer number of FLOPs is required for the compressed layer. For typical compression scenarios where $r \ll \min(m,n)$ and $m \approx n$, this ratio approaches $\frac{r}{n}$, demonstrating a reduction in FLOPs of the low-rank compressed layer. 

For layers with minimal compression, for example only 5\% compression, low-rank decomposition can provide limited gains for that layer. For instance, parallelization inefficiency from requiring two separate matrix multiplications instead of one for very similar FLOPs. It would require additional data movement between HBM and SRAM. That being said, FLOP improvements can be better realised through efficient fused kernel implementations for low-rank linear projections.

\section{Implementation of ARS}
\label{sec:ars_caveats}
Our benchmarking of ARS \citep{adaptive} initially used a slight modification of the original approach, in which the full hypernetwork—including the GRU module—was instantiated separately for each layer to be compressed. In response to feedback, we later aligned our implementation with the original paper by using a single GRU hypernetwork shared across all layers, with only the linear projection layers instantiated per compressed layer. This change resulted in negligible differences in performance metrics; moreover, the initial implementation yielded slightly better results. Therefore, for fairness and competitiveness, we report the best-performing configuration. The full implementation has been open-sourced \href{https://github.com/sidhantls/adaptive-rank-selection-svd}{here}. 

\clearpage
\onecolumn
\section{More LLM Pruner Benchmarks}
\label{sec:llm_pruner_llama3}

In addition to the models reported in \cref{tab:combined_metrics}, we also evaluate performance on Llama-3-8B. Because the current open-source implementation of SVD-LLM does not support Llama-3-8B, we include another comparison between our approach and LLM-Pruner only. The results are presented in \cref{tab:combined_metrics_llama3}.

\begin{table*}[h!]
\centering
\setlength\tabcolsep{2pt}
\small
\resizebox{\textwidth}{!}{
\begin{NiceTabular}{lccccccccccc}
\toprule
\textbf{Method} & \textbf{Param Ratio} & \multicolumn{5}{c}{\textbf{LLaMA-2-7b}} & \multicolumn{5}{c}{\textbf{LLaMA-3-8b}} \\
\cmidrule(lr){3-7} \cmidrule(lr){8-12}
 &  & \textbf{NQ-Open} & \textbf{MMLU} & \textbf{BoolQ} & \textbf{PIQA} & \textbf{OQA} & \textbf{NQ-Open} & \textbf{MMLU} & \textbf{BoolQ} & \textbf{PIQA} & \textbf{OQA} \\
\midrule
Baseline & 1.00 & 26.0 & 41.3 & 77.8 & 78.1 & 31.4 & 29.0 & 62.1 & 81.3 & 79.7 & 34.8 \\
\midrule
Pruner & 0.90 & 15.5 & 28.5 & 64.8 & 77.3 & 31.0 & 17.0 & 41.7 & 68.4 & \underline{77.9} & 31.0 \\
Ours & 0.90 & \underline{22.1} & \underline{37.8} & \underline{77.3} & \underline{78.1} & \underline{33.4} & \underline{19.6} & \underline{44.8} & \underline{71.9} & 77.7 & \underline{32.4} \\
\midrule
Pruner & 0.85 & 13.1 & 24.8 & 67.9 & \underline{77.5} & 31.0 & 11.8 & \underline{35.3} & 56.6 & \underline{77.4} & 28.8 \\
Ours & 0.85 & \underline{18.5} & \underline{33.2} & \underline{74.8} & 77.2 & \underline{32.4} & \underline{14.0} & 29.4 & \underline{61.4} & 75.5 & \underline{29.6} \\
\midrule
Pruner & 0.80 & 7.8 & 25.1 & 64.6 & \underline{76.2} & 29.0 & 5.5 & 23.0 & 53.5 & \underline{74.2} & 24.4 \\
Ours & 0.80 & \underline{13.5} & \underline{29.3} & \underline{71.4} & 75.1 & \underline{32.8} & \underline{8.6} & \underline{24.8} & 53.5 & 74.1 & \underline{25.4} \\
\midrule
\multicolumn{11}{l}{LLM Pruner performance with additional fine-tuning on Alpaca dataset}\\
\midrule
{Pruner+Finetune} & 0.90 & 17.9 & 34.0 & 71.4 & 78.1 & 33.0 & 19.0 & 48.0 & 76.1 & 79.5 & 35.0 \\
{Pruner+Finetune} & 0.85 & 14.7 & 31.4 & 72.2 & 78.7 & 33.0 & 14.7 & 42.9 & 74.4 & 79.0 & 30.8 \\
{Pruner+Finetune} & 0.80 & 11.1 & 26.7 & 67.8 & 77.8 & 30.4 & 10.6 & 32.9 & 70.2 & 78.2 & 29.8 \\
\bottomrule
\end{NiceTabular}
}
\caption{Performance comparison between LLM-Pruner and Our approach on LLAMA-2-7B and LLAMA-3-8B}
\label{tab:combined_metrics_llama3}
\end{table*}

\section{Tabulated Performance of Rank-Selection Models}
\label{sec:all_metrics_appendix}

\subsection{Performance on Llama-2-7b/Llama-3-8b}
\label{sec:llama2_appendix}
\begin{table}[htbp]
\centering
\resizebox{\textwidth}{!}{%
\setlength\tabcolsep{3pt} %
\renewcommand{\arraystretch}{1.1} %
\begin{NiceTabular}{lccccccccccc}
\toprule
\textbf{Method} & \textbf{Param Ratio} & \multicolumn{5}{c}{\textbf{LLaMA-2-7b}} & \multicolumn{5}{c}{\textbf{LLaMA-3-8b}} \\
\cmidrule(lr){3-7} \cmidrule(lr){8-12}
 &  & \textbf{NQOpen} & \textbf{MMLU} & \textbf{BoolQ} & \textbf{PIQA} & \textbf{OQA} & \textbf{NQOpen} & \textbf{MMLU} & \textbf{BoolQ} & \textbf{PIQA} & \textbf{OQA} \\
\midrule
Baseline & 1.00 & 26.0 & 41.3 & 77.8 & 78.1 & 31.4 & 29.0 & 62.1 & 81.3 & 79.7 & 34.8 \\
\midrule
Fixed Rate & 0.90 & 3.41 & 26.3 & 56.3 & 67.7 & 23.2 & 3.10 & 37.6 & \underline{74.6} & 70.1 & 23.2 \\
STRS & 0.90 & 21.2 & 37.6 & 75.8 & 77.5 & 31.8 & 17.7 & \underline{49.7} & 73.2 & \underline{77.7} & 31.4 \\
ARS & 0.90 & 8.01 & 29.0 & 51.9 & 72.2 & 25.4 & 9.09 & 34.8 & 66.6 & 73.9 & 27.6 \\
\textbf{Ours} & 0.90 & \underline{22.1} & \underline{37.8} & \underline{77.3} & \underline{78.1} & \underline{33.4} & \underline{19.6} & 44.8 & 71.9 & 77.7 & \underline{32.4} \\
\midrule%
Fixed Rate & 0.85 & 1.94 & 23.7 & 49.9 & 65.2 & 21.2 & 1.69 & 29.0 & 64.6 & 65.2 & 19.6 \\
STRS & 0.85 & 16.3 & 32.4 & \underline{75.1} & 76.0 & 31.4 & 4.76 & \underline{33.0} & \underline{68.6} & 70.2 & 22.8 \\
ARS & 0.85 & 4.40 & 23.1 & 51.6 & 69.9 & 23.0 & 0.17 & 22.9 & 39.3 & 60.8 & 16.5 \\
\textbf{Ours} & 0.85 & \underline{18.5} & \underline{33.2} & 74.8 & \underline{77.2} & \underline{32.4} & \underline{14.0} & 29.4 & 61.4 & \underline{75.5} & \underline{29.6} \\
\midrule %
Fixed Rate & 0.80 & 0.53 & 23.7 & 60.0 & 62.9 & 20.8 & 0.33 & \underline{24.9} & \underline{63.0} & 60.9 & 17.4 \\
STRS & 0.80 & 9.25 & 28.1 & \underline{71.6} & 71.2 & 28.2 & 0.25 & 24.9 & 49.1 & 63.3 & 16.6 \\
ARS & 0.80 & 0.89 & 23.0 & 59.1 & 65.3 & 19.8 & 0.17 & 24.0 & 60.0 & 62.9 & 18.2 \\
\textbf{Ours} & 0.80 & \underline{13.5} & \underline{29.3} & 71.4 & \underline{75.1} & \underline{32.8} & \underline{8.59} & 24.8 & 53.5 & \underline{74.1} & \underline{25.4} \\
\bottomrule
\end{NiceTabular}
}
\caption{Comparison of evaluation performance using different rank selection methods on LLaMA-2-7b and LLaMA-3-8b. We compare Fixed Rate (naive baseline), STRS \cite{asvd}, ARS \cite{adaptive}, and our proposed approach.}
\label{tab:combined_benchmark}
\end{table}

\subsection{Performance on Gemma-7B}
\label{sec:gemma_appendix}
This section contains the table of metrics of Gemma-7B. 

\begin{table}[htbp]
\centering

\begin{tabular}{lcccccc}
\toprule
\textbf{Method} & \textbf{Param Ratio} & \textbf{NQ-Open} & \textbf{MMLU} & \textbf{BoolQ} & \textbf{PIQA} & \textbf{OpenbookQA} \\ 
\midrule
Original & 1.00 & 25.7 & 62.0 & 83.4 & 80.1 & 32.4 \\ 
\midrule
Fixed Rate   & 0.90 & 2.88 & 33.8 & 71.3 & 67.8 & 26.6 \\ 
STRS     & 0.90 & 14.3 & 45.4 & 81.5 & 78.2 & 31.4 \\ 
ARS     & 0.90 & 15.3 & 29.4 & 67.2 & 76.7 & 30.2 \\ 
\textbf{Ours}     & 0.90 & 13.2 & 49.6 & 77.2 & 76.1 & 32.2 \\
\midrule
Fixed Rate & 0.85 & 0.39 & 27.4 & 62.6 & 59.4 & 19.4 \\
STRS     & 0.85 & 2.30 & 29.3 & 70.0 & 64.4 & 27.6 \\ 
ARS     & 0.85 & 7.40 & 24.3 & 60.7 & 73.7 & 27.4 \\
\textbf{Ours}     & 0.85 & 11.3 & 37.0 & 78.3 & 75.1 & 34.4 \\
\midrule
Fixed Rate & 0.80 & 0.14 & 23.0 & 56.8 & 57.0 & 18.8 \\
STRS     & 0.80 & 0.11 & 23.4 & 51.2 & 60.0 & 21.6 \\ 
ARS     & 0.80 &  2.80 & 23.1 & 54.0 & 71.3 & 27.2 \\ 
\textbf{Ours}     & 0.80 & 8.39 & 33.6 & 75.8 & 74.7 & 34.6 \\ 
\bottomrule 
\end{tabular}
\caption{Evaluation performance of Gemma-8B using different rank selection methods}
\end{table}

\subsection{Performance on Llama-2-13B}
\label{sec:llama13b_appendix}
\begin{table}[htbp]
\centering
\begin{tabular}{lcccccc}
\toprule
Method & Param Ratio & NQOpen & MMLU & BoolQ & PIQA & OpenbookQA \\
\midrule
Original & 1.00 & 30.7 & 52.1 & 80.1 & 79.1 & 35.2 \\ 
\midrule
Fixed & 0.90 & 13.2 & 41.2 & 79.6 & 75.0 & 32.0 \\
STRS & 0.90 & \underline{27.6} & 49.1 & 80.6 & 78.0 & \underline{35.8}\\
ARS & 0.90 & 11.1 & 39.9 & 74.0 & 75.8 & 30.2 \\
\textbf{Ours} & 0.90 & 26.5 & \underline{49.7} & \underline{81.1} & \underline{78.6} & 35.4 \\ 
\midrule
Fixed & 0.85 & 10.1 & 36.1 & 77.9 & 73.6 & 29.6 \\
STRS & 0.85 & \underline{23.7} & 44.9 & 78.8 & \underline{78.3} & 35.2 \\
ARS & 0.85 & 7.22 & 38.3 & 71.3 & 72.0 & 23.8 \\
\textbf{Ours} & 0.85 & 22.7 & \underline{48.1} & \underline{80.5} & 77.4 & \underline{36.0} \\ 
\midrule
Fixed & 0.80 & 6.4 & 36.3 & 77.7 & 71.5 & 27.8 \\
STRS & 0.80 & 17.5 & 32.1 & 77.8 & 75.6 & 31.6 \\
ARS & 0.80 & 3.13 & 24.7 & 63.2 & 67.6 & 23.8 \\
\textbf{Ours} & 0.80 & \underline{18.9} & \underline{44.1} & \underline{81.3} & \underline{77.9} & \underline{36.0} \\
\bottomrule
\end{tabular}
\caption{Evaluation performance of Llama-2-13B using different rank selection methods}
\end{table}

\clearpage
\twocolumn

\section{More Ablations}
\subsection{Albation: Pre-Training Objective}
\label{sec:pretrain_appendix}
\subsubsection{Pre-Training Objective Ablation}
\begin{figure}[h!]
  \centering
  \includegraphics[width=0.4\textwidth,clip, trim=.5em 0em .5em 0em]{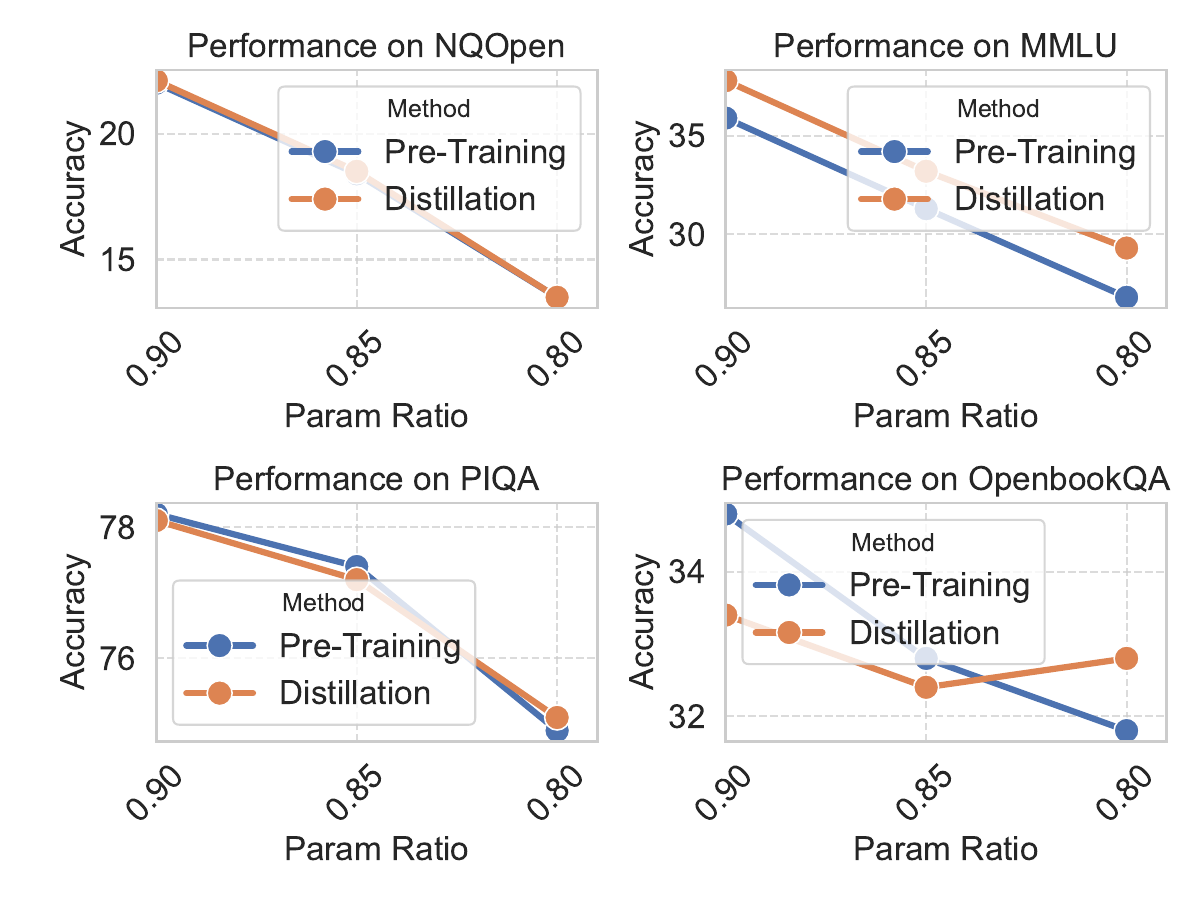}
  \vspace{-1em}
  \caption{Comparison of performance of models trained using distillation objective and a pre-training objective (next word prediction)}
  \label{fig:pretrain}
\end{figure}

In our final results, we utilised a distillation objective that minimised the divergence between the activations of the compressed model and the original model. While both methods lead to very similar results, as shown in \cref{fig:pretrain}, indicating the flexibility of our masking training procedure, learning the optimal ranks through distillation leads to better generalisation on the majority of datasets. At 20\% compression, distillation outperforms pre-training in all datasets. 

For the pre-training objective, the same dataset was used and documents were packed to a sequence length of 256 tokens.

\subsubsection{Pre-Training Objective Hyper-Parameters}
The hyperparameters used in the training process for the pre-training objective are summarised in this section. We sampled 70,000 documents from the Wikitext dataset \citep{wikitext}, which was the same data source used for the model trained on the distillation objective. We utilised document packing, ensuring that all documents contained 256 tokens. Besides this, all parameters, such as optimiser, maximum number of tokens, learning rate, and early stopping criteria, were the same as the experiment that used distillation, as mentioned in Section \ref{distillation_exp}.

\section{Bibliographical References}\label{sec:reference}

\bibliographystyle{lrec2026-natbib}
\bibliography{references}

\end{document}